\documentclass[10pt, a4paper]{article}
\usepackage{lrec2022} % this is the new LREC2022 Style
\usepackage{multibib}
\newcites{languageresource}{Language Resources}
\usepackage{graphicx}
\usepackage{tabularx}
\usepackage{soul}
\usepackage{titlesec}
\titleformat{\section}{\normalfont\large\bfseries\center}{\thesection.}{1em}{}
\titleformat{\subsection}{\normalfont\SmallTitleFont\bfseries\raggedright}{\thesubsection.}{1em}{}
\titleformat{\subsubsection}{\normalfont\normalsize\bfseries\raggedright}{\thesubsubsection.}{1em}{}
\renewcommand\thesection{\arabic{section}}
\renewcommand\thesubsection{\thesection.\arabic{subsection}}
\renewcommand\thesubsubsection{\thesubsection.\arabic{subsubsection}}
\usepackage{epstopdf}
\usepackage[utf8]{inputenc}

\usepackage{hyperref}
\usepackage{xstring}

\usepackage{color}
\usepackage{amssymb}
\usepackage{amsmath}
\usepackage{multirow}
\DeclareMathOperator*{\argmax}{argmax}
\usepackage{makecell}

\usepackage{bera}% optional: just to have a nice mono-spaced font
\usepackage{listings}
\usepackage{xcolor}

\colorlet{punct}{red!60!black}
\definecolor{background}{HTML}{EEEEEE}
\definecolor{delim}{RGB}{20,105,176}
\colorlet{numb}{magenta!60!black}

\lstdefinelanguage{json}{
    basicstyle=\normalfont\ttfamily,
    showstringspaces=false,
    breaklines=true,
    frame=lines,
    backgroundcolor=\color{background},
    literate=
     *{:}{{{\color{punct}{:}}}}{1}
      {,}{{{\color{punct}{,}}}}{1}
      {\{}{{{\color{delim}{\{}}}}{1}
      {\}}{{{\color{delim}{\}}}}}{1}
      {[}{{{\color{delim}{[}}}}{1}
      {]}{{{\color{delim}{]}}}}{1},
}

\title{MeSHup: A Corpus for Full Text Biomedical Document Indexing}

\name{Xindi Wang$^{1,3}$, Robert E. Mercer$^{1,3}$, Frank Rudzicz$^{2,3,4}$} 

\address{$^1$Department of Computer Science, University of Western Ontario, London, Ontario, Canada\\
$^2$Department of Computer Science, University of Toronto, Toronto, Ontario, Canada \\
$^3$Vector Institute, Toronto, Ontario, Canada\\
$^4$ Unity Health Toronto, Toronto, Ontario, Canada\\
         xwang842@uwo.ca, mercer@csd.uwo.ca, frank@cs.toronto.edu}

\abstract{
Medical Subject Heading (MeSH) indexing refers to the problem of assigning a given biomedical document with the most relevant labels from an extremely large set of MeSH terms. Currently, the vast number of biomedical articles in the PubMed database are manually annotated by human curators, which is time consuming and costly; therefore, a computational system that can assist the indexing is highly valuable. When developing supervised MeSH indexing systems, the availability of a large-scale annotated text corpus is desirable. A publicly available, large corpus that permits robust  evaluation and comparison of various systems is important to the research community. We release a large scale annotated MeSH indexing corpus, MeSHup, which contains 1,342,667 full text articles in English, together with the associated MeSH labels and metadata, authors, and publication venues that are collected from the MEDLINE database. We train an end-to-end model that combines features from  documents  and their associated labels on our corpus and report the new baseline. 
 \\ \newline  }

\begin{document}

\maketitleabstract

\section{Introduction}
MEDLINE\footnote{\urlstyle{same}\url{https://www.nlm.nih.gov/medline/medline_overview.html}} comprises 33 million (as of Nov. 2021) references to journal articles in the life sciences with a concentration on biomedicine, which is the National Library of Medicine’s\footnote{\urlstyle{same}\url{https://www.nlm.nih.gov}} (NLM) premier bibliographic database. It includes textual information (title and abstract) and bibliographic information for articles from academic journals covering various disciplines of the life sciences and biomedicine. PubMed\footnote{\urlstyle{same}\url{https://pubmed.ncbi.nlm.nih.gov/about/}} is a free search engine that provides free access to the MEDLINE database. In addition to MEDLINE, PubMed also provides access to the PubMed Central\footnote{\urlstyle{same}\url{https://en.wikipedia.org/wiki/PubMed\_Central}} (PMC) repository that archives open-access, full-text scholarly articles in biomedical and life sciences journals. All records in the MEDLINE database are indexed with \textbf{Me}dical \textbf{S}ubject \textbf{H}eadings (MeSH)\footnote{\urlstyle{same}\url{https://www.nlm.nih.gov/mesh/meshhome.html}} -- a controlled and hierarchically-organized vocabulary produced and maintained by the NLM. As of 2021, there are 29,369 main MeSH headings, and each citation is indexed with 13 MeSH terms on average. MeSH headings can be further qualified by 83 subheadings (also known as qualifiers). In addition, Supplementary Concept Records (SCRs) refer to specific chemical substances in the MEDLINE records.

MeSH indexing, a process that annotates documents with  concepts from established semantic taxonomies and ontologies, is important for biomedical text classification and information retrieval. MEDLINE citations are indexed by human annotators who read the full text of the article and assign the most relevant MeSH labels to the articles. The manual annotation process ensures the high quality of indexing but, inevitably, the cost can be prohibitive. There has been a steady and sizeable increase in the number of citations that are added to the MEDLINE database every year; for instance, in 2020, 952,919 articles were added (approximately 2,600 on a daily basis)\footnote{\urlstyle{same}\url{https://www.nlm.nih.gov/bsd/medline_pubmed_production_stats.html}}and the average cost of annotation per document is approximately \$9.40 \cite{Mork2013TheNM}. Faced with the growing workload, NLM annotators remain tasked with indexing newly published articles efficiently and promptly. 
Therefore, there is a pressing need for automatic supports to indexing biomedical literature. 

Many state-of-the-art models have been proposed to deal with MeSH indexing; however, there is a clear drawback to these automatic indexing systems because of the data used to train them. Existing corpora for MeSH indexing only provide the title and abstract, while human annotators review full text articles. This suggests that important information might be contained in the full text which is available to the human indexer, but does not appear in the title and abstract that automatic indexing systems use to make recommendations. \newcite{Mork201712YO} further indicated that some sections in the full text that include very specific information, such as the ``Methods'' section, help improve the performance of automatic models. Thus, a corpus that contains full text articles and their associated MeSH labels is highly desirable.

In this work, we construct a new labeled full text MeSH indexing dataset\footnote{\urlstyle{same}\url{https://github.com/xdwang0726/MeSHup}}, MeSHup, that for each entry, is a mashup of the PMID, title, abstract, journal, year, author list, MeSH terms, chemical list, and Supplementary Concept Records from MEDLINE and the full text introduction, methods, results, discussion, figure captions, and table captions that are available from BioC-PMC \cite{Comeau2019}. %\citelanguageresource{10.1093/bioinformatics/btz070}. 
To the best of our knowledge, MeSHup is the first publicly available (and the largest) full text dataset annotated for MeSH indexing. We also propose a multi-channel model that incorporates extracted features from different sections of the full text and report its baseline results. 

\section{Related Work}
\subsection{Biomedical Corpora Related to MeSH Terms}
There are several corpora in the biomedical domain that contain or make use of MeSH terms. The OSHUMED test collection \citelanguageresource{Hersh1994OHSUMEDAI} is a set of 348,566 clinically-oriented references in the MEDLINE database which are obtained from 270 medical journals in the years 1987 to 1991. For each citation, the collection contains the title, abstract, MeSH indexing terms, author, source, and publication type. The OSHUMED corpus is one of the earliest corpora that is related to the MeSH indexing task. The GENIA corpus \citelanguageresource{Kim2003GENIAC} is a valuable resource in the biomedical literature that was created to support the development of tools for text mining and information retrieval and their evaluation. It contains 2,000 abstracts taken from the MEDLINE database with a variety of entity types in the GENIA Chemical ontology that are derived from MeSH terms. The CHEMDNER corpus \citelanguageresource{Krallinger2015TheCC} contains 10,000 PubMed abstracts and 84,355 manually annotated chemical entities. CHEMDNER labels entities based on the Chemicals and Drugs branch of the MeSH hierarchy and the MeSH substances. 
The BioCreative V Chemical-Disease Relation Task Corpus (BC5CDR) \cite{BC5CDR}
was developed for the BioCreative V challenge. A team of Medical Subject Headings (MeSH) indexers for disease/chemical entity annotation and Comparative Toxicogenomics Database (CTD) curators for CID relation annotation were invited to ensure high annotation quality and productivity. Detailed annotation guidelines and automatic annotation tools were provided. The resulting corpus consists of 1500 PubMed articles with 4409 annotated chemicals, 5818 diseases and 3116 chemical-disease interactions. Each entity annotation includes both the mention text spans and normalized concept identifiers, using MeSH as the controlled vocabulary.
The NLM-CHEM corpus \cite{Islamaj2021NLM-Chem}, created for Track 2 of BioCreative VII, consists of 150 full text articles with chemical entity annotations provided by human experts for ~5000 unique chemical names, mapped to ~2000 MeSH identifiers. This dataset is compatible with the CHEMDNER and BC5CDR corpora described above.

\subsection{Related Full-Text Biomedical Corpora}
A number of full-text corpora have been created in the biomedical domain. Each has been generated for specific purposes and consequently has been annotated with that in mind.
The earliest is a small five biomedical paper corpus \cite{Gasperin2007} which was designed to capture anaphora. The corpus of biomedical articles are annotated with anaphoric links between coreferent and associative (biotype, homolog, and set-member) noun phrases referring to the biomedical entities of interest to the authors.
For the BioCreative I task 2 \cite{Hirschman2005Biocreative}, the training set corpus comprised 803 full text articles from four different journals that had been previously annotated for human protein function and the test set corpus comprised 212 full text articles.
The CRAFT (The Colorado Richly Annotated Full-Text) corpus \cite{Verspoor2012} is a collection of 97 articles from the PubMed Central Open Access subset. Each article has been manually annotated along structural, coreference, and concept dimensions.
More recently, the BioC-BioGRID corpus \cite{BioC-BioGRID} comprises full text articles annotated for protein-protein and genetic interactions.

\subsection{Automatic MeSH Indexing Based on Title and Abstract}
As discussed, there is  rapid growth in the number of articles in MEDLINE, and the NLM has developed an indexing tool, Medical Text Indexer (MTI), to recommend MeSH terms in automated and semi-automated modes \cite{Aronson2004TheNI}. MTI first takes the title and abstract of the input article and generates recommended MeSH terms, using a ranking algorithm to determine final suggestions. There are two  important components in MTI: MetaMap Indexing (MMI) and PubMed-Related Citations (PRC) \cite{Lin2007,10.1136/jamia.2009.002733}. MetaMap recommends MeSH terms based on the mapping of biomedical concepts in the title and abstract of the input article to the the Unified Medical Language System\footnote{\urlstyle{same}\url{https://www.nlm.nih.gov/research/umls/}} (UMLS). PRC suggests MeSH terms by looking at similar annotations in MEDLINE using $k$-nearest neighbours. Two sets of recommended MeSH terms are combined to generate the final MeSH list. 

Since 2013, BioASQ\footnote{\urlstyle{same}\url{http://bioasq.org}}\cite{283} has organized the biomedical semantic indexing challenge, which offers the opportunity for more participants to get involved in the MeSH indexing task. BioASQ provides annotated PubMed articles with the title and abstract only, and participants can tune their annotation models accordingly. Many effective indexing systems have been proposed since then, such as MeSHLabeler \cite{Liu2015MeSHLabelerIT}, DeepMeSH \cite{Peng2016DeepMeSHDS}, AttentionMeSH \cite{Indexer2018AttentionMeSHS}, MeSHProbeNet \cite{Xun2019MeSHProbeNetAS}, and KenMeSH \cite{Wang2022KenMeSHKE}. MeSHLabeler and DeepMeSH are models based on a Learning-to-Rank (LTR) framework. AttentionMeSH and MeSHProbeNet both utilize deep recursive neural networks (RNNs) and attention mechanisms, where the main difference is that the former uses label-wise attention while the latter employs multi-view self-attentive MeSH probes. KenMeSH combines text features and the MeSH label hierarchy by using a dynamic knowledge-enhanced mask to index MeSH terms.

\subsection{MeSH Indexing Based on Full Text}
MeSH indexing with full texts has been studied using relatively small sets of data or restricted to small numbers of specific MeSH terms because of the limitation of full text access. \newcite{Gay2005SemiAutomaticIO} collected 500 articles in 17 journal issues in the PubMed database and used the full text as input to MTI. They found that using the full text of an article provides significantly better (7.4\%) quality of automatic indexing than using only abstracts and titles. \newcite{JimenoYepes2012MeSHIB} used a collection of 1,413 biomedical articles randomly selected from the PMC Open Access Subset. They first ran automatic summaries over the full test and then used the generated summary as input to MTI. The experimental results showed that incorporating full texts achieved higher (6\%) recall with a trade-off in precision compared to using the abstracts and titles only. \newcite{DemnerFushman2015ExtractingCO} collected 14,829 citations and used a rule-based method to classify Check Tags, a small set of MeSH terms (29 MeSH Check Tags) that represent characteristics of the subjects. \newcite{Wang2019IncorporatingFC} released a full text dataset curated from PMC with 257,590 articles and employed a multi-channel CNN-based feature extraction model. FullMeSH \cite{FullMeSH} and BERTMeSH \cite{BERTMeSH} used 1.4M full text articles, the former with an attention-based CNN and the latter with pre-trained contextual embeddings together with an attention mechanism. Unfortunately, they did not make their dataset available, which makes it difficult for others to compare and evaluate their work.  
\section{Dataset Construction}
In this subsection, we introduce how to construct the dataset based on the PubMed Central Open Access in BioC format\footnote{\urlstyle{same}\url{https://www.ncbi.nlm.nih.gov/research/bionlp/APIs/BioC-PMC/}} (BioC-PMC) \citelanguageresource{10.1093/bioinformatics/btz070} and the MEDLINE/PubMed Annual Baseline Repository\footnote{\urlstyle{same}\url{https://lhncbc.nlm.nih.gov/ii/information/MBR.html}} (MBR). We download the entire BioC-PMC subset (as of Nov.\ 2021) and obtain 3,601,092 full text articles. We also download the entire MBR collection (as of Nov.\ 2021) and obtain 31,850,051 citations with metadata in the MEDLINE database. In order to reduce bias, we only consider articles indexed by human annotators (i.e., articles in the MEDLINE database with modes marked as `curated' (MeSH terms were provided algorithmically and were human reviewed) or `auto' (MeSH terms were provided algorithmically) are not considered)\footnote{\urlstyle{same}\url{https://www.nlm.nih.gov/bsd/licensee/elements_descriptions.html}}, and we only focus on articles written in English (i.e., only articles annotated as `eng' are considered) in the MEDLINE database. We then match BioC-PMC articles with MBR citations using the PubMed ID (PMID) and obtain a set of 1,342,667 biomedical documents. 
\paragraph{Information extracted from BioC-PMC.} Each article in the BioC-PMC subset is structured in a single XML file. The original published articles are formatted in various ways depending on the publisher. With the BioC format \cite{Comeau2019}, an article's textual information is preserved and each article is organized in a unified structure. We parse the tags in the BioC formatted XML files to get the section names and their corresponding texts. We then divide and normalize all full text articles into eight BioC sections: title, abstract, introduction, methods, results, discussion, figure captions, and table captions. Table \ref{table:1} summarizes the statistical information for the described sections.
\begin{table}[!ht] 
\begin{center}
\resizebox{\columnwidth}{!}{
\begin{tabular}{| c | c | c |}

      \hline
      Sections& number of articles & average length\\
      \hline
      Title & 1,342,667 & 16 \\
      \hline
      Abstract & 1,342,667 & 258\\
      \hline
      Introduction & 1,279,276 & 991\\
      \hline
      Methods & 1,135,757 & 1446\\
      \hline
      Results & 1,090,981 & 1640 \\
      \hline
      Discuss & 1,042,379 & 1249 \\
      \hline
      Figure Captions & 1,155,208 & 560 \\
      \hline
      Table Captions & 520,780 & 123 \\
      \hline

\end{tabular}}
\caption{Statistics of the generated dataset for each of the eight sections} \label{table:1}
\end{center}
\end{table} 
\paragraph{Information extracted from the MEDLINE/PubMed Annual Baseline (MBR).} Starting in 2002, MBR has provided access to all MEDLINE citations and it updates and adds new citations every year. Each year's baseline contains textual information (title and abstract) of the citations as well as various types of metadata, such as their authors, publishing venues, and references. Metadata can be regarded as strong indicators for the semantic indexing task as they include latent information of research topics. We therefore extract the metadata of each article from MBR; the detailed metadata and their descriptions are stated in Table \ref{table:2}.
\begin{table}[t] 
\begin{center}
\resizebox{\columnwidth}{!}{
\begin{tabular}{| c | l |}

      \hline
      Metadata & Descriptions\\
      \hline
      PMID & \makecell[l]{The PubMed (NLM database that\\ incorporates MEDLINE) unique identifier.}\\
      \hline
      Authors & \makecell[l]{Personal and collective (corporate)\\ author names published with the article.}\\
      \hline
      Journal & The journal that the article is published in.\\
      \hline
      Year & The year in which the article is published.\\
      \hline
      DOI &  Digital Object Identifiers.\\
      \hline
      MeSH Terms & \makecell[l]{NLM controlled vocabulary,\\ Medical Subject Headings (MeSH).}\\
      \hline
      Supply MeSH & Supplementary Concept Record (SCR) terms.\\
      \hline
      Chemical List & A list of chemical substances and enzymes.\\
      \hline

\end{tabular}}
\caption{Meta-data extracted from MBR and their descriptions} \label{table:2}
\end{center}
\end{table}

We combine the information extracted from BioC-PMC and MBR to form MeSHup, a new large-scale, full-text biomedical semantic indexing dataset. Specifically, each article in the dataset contains the full textual information and the metadata associated with it. %An extract from the MeSHup is shown in JSON format as follows, and a complete version is in Appendix.
The keys (section descriptors) from MeSHup are shown in JSON format in Fig.\ \ref{JSONkeys}, and an abbreviated version of the dataset is provided in the Appendix. % \ref{sec:appendix}.
\begin{figure}[t]
\begin{lstlisting}[language=json]
{"articles":[
    {"PMID": ,
     "TITLE": ,
     "ABSTRACT": ,
     "INTRO": , 
     "METHODS": ,
     "RESULTS": ,
     "DISCUSS": ,
     "FIG_CAPTIONS": ,
     "TABLE_CAPTIONS": ,
     "JOURNAL": ,
     "YEAR": ,
     "DOI": ,
     "AUTHORS": ,
     "MeSH": ,
     "CHEMICALS": ,
     "SUPPLMeSH":
    },
    {
     ...
    },
    ...
]}
\end{lstlisting}
\caption{The section descriptors given as JSON keys.}\label{JSONkeys}
\end{figure}
Our goal in releasing MeSHup is to promote large-scale ontological classification of biomedical documents, using full texts, across the community. With MeSHup, researchers can explore and test state-of-the-art indexing systems with a common standard.  

\section{Experiments} 
Given the full text of a biomedical article, MeSH indexing can be regarded as a multi-label text classification problem. The learning framework is defined as follows. $\mathcal{X}= \{x_{1}, x_{2}, ..., x_{n}\}$ is a set of biomedical documents and $\mathcal{Y} = \{y_{1}, y_{2}, ..., y_{L}\}$ is the set of MeSH terms. Multi-label classification studies the learning function $f: \mathcal{X}  \rightarrow [0, 1]^{\mathcal{Y}}$ using the training set $\mathcal{D} = \{(x_{i}, Y_{i}), i = 1, ..., n\}$, where $n$ is the number of documents in the set, and $Y_{i} \subset \mathcal{Y}$ is the set of MeSH labels for document $x_{i}$. The objective of MeSH indexing is to predict the correct MeSH labels for any unseen article $x_{k}$, where $x_{k}$ is not in $\mathcal{X}$.

\subsection{Baseline Model}
\begin{figure}[!ht]
\begin{center}
\includegraphics[width=\columnwidth]{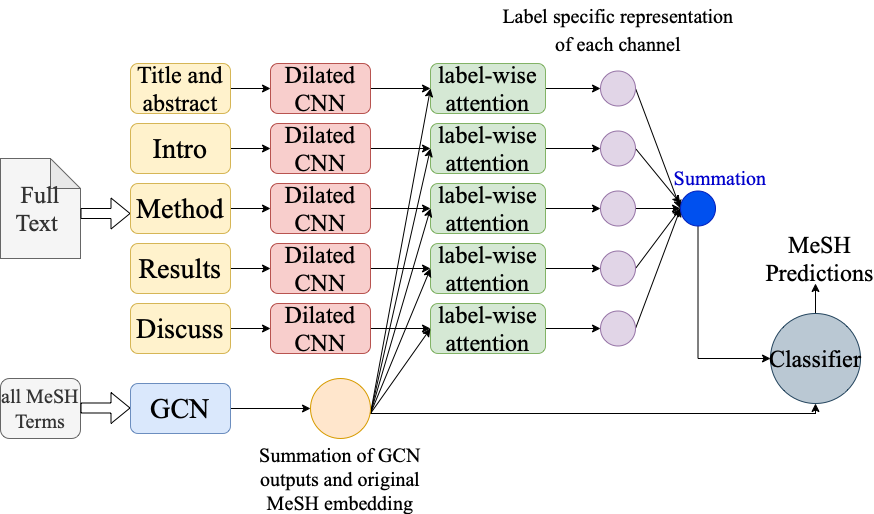}
\caption{Model Architecture - There are three main components in our method. First, a multi-channel document representation module operates on each section of an input article. Second, a 2-layer GCN creates label vectors. Lastly, a label-wise attention component calculates the label-specific attention vectors that are used for predictions.}\label{fig:1}
\end{center}
\end{figure}

In general, traditional MeSH indexing systems focus on the document features only, thereby suffering from a lack of information about the MeSH hierarchy. To handle this, we present a hybrid document-label feature method, which is composed of a multi-channel document representation module, a label feature representation module, and a classifier. The overall model architecture is shown is Figure \ref{fig:1}. 
\subsubsection{Multi-channel Document Representation Module}

The multi-channel document representation module has five input channels: the title and abstract channel, the introduction channel, the methods channel, the results channel, and the discussion channel. Texts in each channel are represented by the embedding matrices, namely $E_{\textit{C}} \in \mathbb{R}^{d}$, where $d$ represents the dimension of the word embeddings. 

Instead of a recurrent model which poses computational and technical issues, we apply a multi-level dilated convolutional neural network (DCNN) to each channel in order to get the distant effect memory from a CNN model. To be specific, our DCNN is a three-layer, one-dimensional convolutional neural net (CNN) with dilated convolution kernels, which obtain high-level semantic representations of texts without increasing the computation. The concept of dilated convolution has been popular in semantic segmentation in computer vision in recent years \cite{Yu2016MultiScaleCA,Li2018CSRNetDC}, and it has been introduced to sequential data \cite{Bai2018ConvolutionalSM}, specially to the field of NLP in neural machine translation \cite{kalchbrenner2017neural} and text classification \cite{lin-etal-2018-semantic-unit}. Dilated convolution enables exponentially large receptive fields over the embedding metric, which  captures long-term dependencies over the input texts. 

We apply a multi-level DCNN with different dilation rates on top of the embedding metric on each channel. Larger dilations represent a wider range of inputs that can capture  sentence-level information, whereas small dilations capture word-level information. The semantic features returned by DCNN for each channel is denoted as $D_{\textit{C}}\in \mathbb{R}^{(l - s + 1) \times 2d}$, where $l$ is the sequence length in channel $C$ and $s$ is the width of the convolution kernels.

\subsubsection{Label Feature Representation Module}

Graph convolutional neural networks (GCNs) \cite{Kipf2017SemiSupervisedCW} have attracted wide attention recently. They have been effective in tasks that have rich relational structures, as GCNs preserve global information within the graph. Traditional multi-label text classification mainly focuses on local consecutive word sequences; for instance, two deep networks commonly used in building text representations after learning word embeddings are convolutional (CNNs) and recurrent neural networks (RNNs)  \cite{kim-2014-convolutional,tai-etal-2015-improved}. Some recent studies explored GCN for text classification, where they either viewed a document or a sentence as a graph of word nodes \cite{Peng2018LargeScaleHT,Yao2019GraphCN}.
\begin{figure}[!ht]
\begin{center}
\includegraphics[width=\columnwidth]{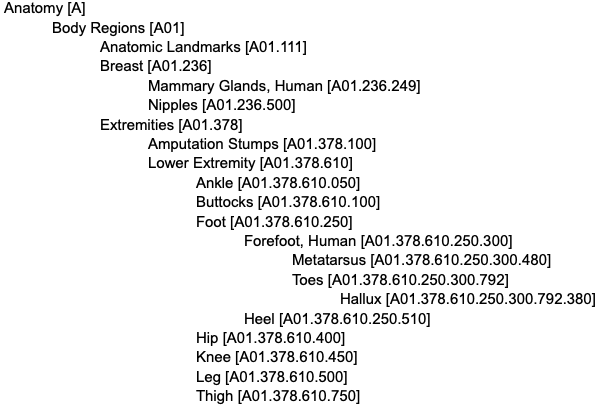}
\caption{An example of the MeSH hierarchy (retrieved Nov.\ 2021). For example, under Body Regions, there are specific body regions under the MeSH tree.}\label{fig:2}
\end{center}
\end{figure} 
MeSH labels are arrayed hierarchically, an example of which is shown in Figure \ref{fig:2}. We take advantage of the structured knowledge we have over the parent and child relationships in the MeSH label space by using a GCN. In the label graph setting, we formulate each MeSH label in $\mathcal{Y}$ as a node in the graph. The edges represent parent and child relationships among the MeSH terms. The edge types of a node contain edges from itself, from its parent, and from its children. We first take the average of the word embeddings of the words in the MeSH label descriptor as the initial label embedding $v_{\textit{i}} \in \mathbb{R}^{d}$:
\begin{equation} \label{eq:1}
%    v_{i} = \frac{1}{m_i}\sum_{j \in m_i}w_{j}, i=1, 2, ..., L,
    v_{i} = \frac{1}{m_i}\sum_{j = 1}^{m_i}w_{j}, i=1, 2, ..., L,
\end{equation}
where $m_i$ is the number of words in the descriptor of label $i$, $w_{\textit{j}}$ is the word embedding of word $j$, and $L$ is the number of labels. The GCN layers capture information about immediate neighbours with one layer of convolution, and information from a larger neighbourhood can be integrated by using a multi-layer stack. We use a two-layer GCN to incorporate the hierarchical information among MeSH terms. At each GCN layer, we aggregate the parent and child nodes for the $i^{th}$ label to form the new label embedding for the next layer:
\begin{equation} 
    h^{l+1} = \sigma(A \cdot h^{l} \cdot W^{l}),
\end{equation}
where $h^{l}$ and $h^{l+1} \in \mathbb{R}^{L \times d}$ indicate the node presentations of the $l^{th}$ and $(l+1)^{th}$ layers, $\sigma(\cdot)$ denotes an activation function, $A$ is the adjacency matrix of the MeSH hierarchical graph, and $W^{l}$ is a layer-specific trainable weight matrix. Next, we sum both the averaged description vector from Equation \ref{eq:1} with the GCN output to form:
\begin{equation} 
    H_{\textit{label}} = v + h^{l+1},
\end{equation}
where $H_{\textit{label}} \in \mathbb{R}^{L \times d}$ is the final label matrix.

\subsubsection{Classifier}

For large multi-label text classification tasks, relevant information for each label might be scattered in various locations of the article. In order to match documents to their corresponding label vectors, we employ label-wise attention following \newcite{Mullenbach2018ExplainablePO}. We generate label-specific representations for each channel:
\begin{equation} 
    \begin{gathered}
        \alpha_{\textit{C}} = \textrm{Softmax}(D_{\textit{C}} \cdot H_{\textit{label}}) \\
        \textit{content}_{\textit{C}} = \alpha_{\textit{C}}^{T} \cdot D_{\textit{C}},
    \end{gathered}
\end{equation}
where $\textit{content}_{\textit{C}} \in \mathbb{R}^{L \times d}$. We then sum up the label-specific content representation for each channel to form the final document representation for each article:
\begin{equation} \label{eq:12}
    D = \sum \textit{content}_{\textit{C}}
\end{equation}
We then generate the predicted score for each MeSH label via:
\begin{equation} \label{eq:13}
    \hat{y_{i}} = \sigma(D \odot H_{\textit{label}}), i = 1, 2, ..., L,
\end{equation}
where $\sigma(\cdot)$ denotes the sigmoid function. Training our proposed method uses binary cross-entropy as the loss function:
\begin{equation} \label{eq:14}
    L =\sum_{i=1}^{L}[-y_{i} \cdot \log(\hat{y_{i}}) -  (1-y_{i}) \cdot \log(1 - \hat{y_{i}}))],
\end{equation}
where $y_{i} \in [0,1]$ is the ground truth of label $i$, and $\hat{y_{i}} \in [0,1]$ denotes the prediction of label $i$ obtained from the proposed model.

\subsection{Evaluation Metrics}
We evaluate performance of MeSH indexing systems using two groups of measurements: bipartition-based evaluation and ranking-based evaluation. Bipartition evaluation is further divided into example-based and label-based metrics. Example-based measures, computed per data point, compute the harmonic mean of standard precision (EBP) and recall (EBR) for each data point. The metrics are defined as:
\begin{equation}
    \textit{EBF} = \frac{2 \times\textit{EBR} \times \textit{EBP}}{\textit{EBR} + \textit{EBP}},
\end{equation}
where
\begin{equation}
    \textit{EBP} = \frac{1}{N}\sum_{i = 1}^{N}\frac{| y_{i} \cap \hat{y_{i}} |}{| \hat{y_{i}} | },
\end{equation}
\begin{equation}
    \textit{EBR}= \frac{1}{N}\sum_{i = 1}^{N}\frac{| y_{i} \cap \hat{y_{i}} |}{| y_{i} | },
\end{equation}
where $y_i$ is the true label set and $ \hat{y_{i}}$ is the predicted label set for instance $i$, and $N$ represents the total number of instances.

We perform label-based evaluation for each label in the label set. The measurements include
Micro-average F-measure (MiF) and Macro-average F-measure (MaF). MiF aggregates the global contributions of all MeSH labels and then calculates the harmonic mean of micro-average precision (MiP) and micro-average recall (MiR), which are heavily influenced by frequent MeSH terms. MaF computes the macro-average precision (MaP) and macro-average recall (MaR) for each label and then averages them, which provides equal weight to each MeSH term. Therefore frequent MeSH terms and infrequent ones are equally important. The aforementioned metrics are defined as follows:
\begin{equation}
    \textit{MiF} =  \frac{2 \times\textit{MiR} \times \textit{MiP}}{\textit{MiR} + \textit{MiP}},
\end{equation}
where 
\begin{equation}
    \textit{MiP} = \frac{\sum_{j=1}^{L}\textit{TP}_{j}}{\sum_{j=1}^{L}\textit{TP}_{j} + \sum_{j=1}^{L}\textit{FP}_{j}},
\end{equation}
\begin{equation}
    \textit{MiR} = \frac{\sum_{j=1}^{L}\textit{TP}_{j}}{\sum_{j=1}^{L}\textit{TP}_{j} + \sum_{j=1}^{L}\textit{FN}_{j}},
\end{equation}

\begin{equation}
    \textit{MaF} =  \frac{2 \times\textit{MaR} \times \textit{MaP}}{\textit{MaR} + \textit{MaP}},
\end{equation}
where 
\begin{equation}
    \textit{MaP} = \frac{1}{L}\sum_{j=1}^{L}\frac{\textit{TP}_{j}}{\textit{TP}_{j} + \textit{FP}_{j}},
\end{equation}
\begin{equation}
   \textit{MaR} = \frac{1}{L}\sum_{j=1}^{L}\frac{\textit{TP}_{j}}{\textit{TP}_{j} + \textit{FN}_{j}},
\end{equation}
where $\textit{TP}_{j}$, $\textit{FP}_{j}$ and $\textit{FN}_{j}$ are the true positives, false positives, and false negatives, respectively, for each label $l_{j}$ in the set of all labels $L$.

Ranking-based evaluation includes precision at \textit{k} (\textit{P@k}) and recall at \textit{k} (\textit{R@k}). \textit{P@k} shows the number of relevant MeSH terms that are suggested in the top-$k$ recommendations of the MeSH indexing system, and \textit{R@k} indicates the proportion of relevant items that are suggested in the top-$k$ recommendations. The metrics are defined as follows:
\begin{equation}
    \textit{P@k} = \frac{1}{k}\sum_{l \in r_{k}\left(\hat{y}\right)}y_{l},
\end{equation}
\begin{equation}
    \textit{R@k} = \frac{1}{| y_{i} |}\sum_{l \in r_{k}\left(\hat{y}\right)}y_{l},
\end{equation}
where $r_{k}$ returns the top-$k$ recommended items. 

Thresholds greatly impact the bipartition-based evaluation metrics. We therefore tuned the threshold $\tau_{i}$ for predicting the $i$-th label, and selected the predicted MeSH term (\textit{MeSH}$_{i}$) whose predicted probability is greater than $\tau_{i}$:
\begin{equation} \label{eq:15}
    \textit{MeSH}_{i} = 
    \begin{cases}
        \hat{y_{i}} \geq \tau_{i},  1 \\
        \hat{y_{i}} < \tau_{i}, 0 
    \end{cases} 
\end{equation}
We used the micro-F optimization algorithm proposed by \newcite{Pal2020MultiLabelTC} to tune the thresholds:
\begin{equation}
    \tau_{i} = \argmax_\mathcal{T} \textit{MiF}(\mathcal{T}),
\end{equation}
where $\mathcal{T}$ represents all possible threshold values for label $i$. 
\subsection{Experiment Settings}

We implement our model using PyTorch \cite{NEURIPS2019_9015}. We use 200-dimensional word embeddings (BioWordVec) that are pretrained on PubMed article titles and abstracts \cite{Zhang2019BioWordVecIB}. For the model's DCNN component, we use a 1-dimension convolution with kernel size 3 and a three-level dilated convolution with dilation rates [1,2,3]. The number of hidden units in both components of our model is set to 200. We use the Adam optimizer \cite{Kingma2015AdamAM} with a minibatch size of 8 and an initial learning rate of 0.0003 with a decay rate of 0.9 in every epoch. To avoid overfitting, we apply dropout directly after the embedding layer with a rate of 0.2 and use early stopping strategies \cite{Yao2007OnES}. Our model is trained on a single NVIDIA A100 GPU. It takes approximately five to seven days to train the full model.

\subsection{Experimental Results}

In the baseline model, we are interested in the articles that have all six sections, i.e., title, abstract, introduction, method, results, and discuss. We extract the 957,426 articles from MeSHup that meet these criteria. We use stratified sampling over publication year to split our dataset into training, validation and testing. We use 80\% of the documents for training (765,920), 10\% for validation (95,737), and 10\% for testing (95,769). 

We first conduct our experiments with titles and abstracts only, and then we do our experiments on the full texts. From this experiment, we would like to see how integrating full text information affects the indexing performance compared with using the titles and abstracts only. Table \ref{table:3} summarizes the results of bipartition evaluation and Table \ref{table:4} shows the results of ranking-based measures. We can see substantial improvements on all evaluation metrics when involving full texts, which indicates that full texts are more informative compared to the titles and abstracts. The baseline model preforms fairly well on precisions, but with a trade off in recalls. The reason for this could be that the frequency of each MeSH label is quite biased, some labels might have very few training examples so that the baseline model is very hard to predict those rare labels. 
\begin{table}[t]
\centering
\resizebox{\columnwidth}{!}{
\begin{tabular}{| c  c | c | c |}
\hline
\multicolumn{2}{|c|}{\textit{Bipartition evaluation}} & 
\multicolumn{2}{|c|}{\textit{Methods}} \\ \hline

&  & \textit{Titles and Abstracts} & \textit{Full Texts}\\
\hline
\multirow{3}{*}{\centering \textit{Example based}}&\textit{EBF} & 0.183 & \textbf{0.259}\\
~ & \textit{EBP} & 0.503 & \textbf{0.588}\\
~ & \textit{EBR} & 0.112 & \textbf{0.166}\\\hline
\multirow{3}{*}{\centering \textit{Micro-averaged}}&\textit{MiF} & 0.177 & \textbf{0.259}\\
~ & \textit{MiP} & 0.473 & \textbf{0.604}\\
~ & \textit{MiR} & 0.110 & \textbf{0.164} \\\hline
\multirow{3}{*}{\centering \textit{Macro-averaged}}&\textit{MaF} & 0.362 & \textbf{0.367}\\
~ & \textit{MaP} & 0.798 & \textbf{0.810} \\
~ & \textit{MaR} & 0.234 & \textbf{0.237} \\\hline
\end{tabular} }
\caption{Comparison using only titles and abstracts and full texts across bipartition evaluation. Bold: best scores in each row.}\label{table:3}
\end{table}

\begin{table}[t]
\centering
\resizebox{\columnwidth}{!}{
\begin{tabular}{| c  c | c | c |}
\hline
\multicolumn{2}{|c|}{\textit{Ranking Based}} & 
\multicolumn{2}{|c|}{\multirow{2}{*}{\textit{Methods}}} \\ %} \\

\multicolumn{2}{|c|}{\textit{Measure}}  & \multicolumn{2}{|c|}{} \\
\hline
&   & \textit{Titles and Abstracts} & \textit{Full Texts}\\
\hline
\multirow{5}{*}{\centering \textit{P@k}}&\textit{$P@1$} & 0.699 & \textbf{0.801}\\
~ & \textit{$P@3$} &  0.462 & \textbf{0.609} \\
~ & \textit{$P@5$} & 0.372 & \textbf{0.496} \\
~ & \textit{$P@10$} & 0.260 & \textbf{0.341}\\
~ & \textit{$P@15$} &  0.205 &  \textbf{0.267}\\\hline
\multirow{5}{*}{\centering \textit{R@k}}&{$R@1$} & 0.051 & \textbf{0.077}\\
~ & \textit{$R@3$} & 0.098 & \textbf{0.128}\\
~ & \textit{$R@5$} & 0.131 & \textbf{0.171}\\
~ & \textit{$R@10$} & 0.180 & \textbf{0.232}\\
~ & \textit{$R@15$} & 0.214 & \textbf{0.272}\\\hline
\end{tabular} }
\caption{Comparison using only titles and abstracts and full texts across ranking-based measures. Bold: best scores in each row.}\label{table:4}
\end{table}

\section{Conclusion and Future Work}
We present MeSHup, a new, publicly available full text dataset annotated for MeSH indexing. It is a mashup of full-text information from BioC-PMC and associated metadata collected from the MEDLINE database. This is the first large dataset that contains the full text information that allows the research community to incorporate more textual information other than the title and abstract in building MeSH indexing systems. We also train an end-to-end model that comprises features extracted from the document itself and features obtained from labels.  

We think that the MeSHup dataset could be a valuable resource not only for MeSH indexing but also for full text mining and retrieval. Since our analysis covers several but not all sections of the full text articles, it is likely that other parts of the article together with metadata may also have impacts on future outcomes. In future, we plan to involve more sections of the textual information and metadata to improve the automatic indexing system. 

\section*{Acknowledgements}
We thank all reviewers for their constructive comments and feedback. Resources used in preparing this research were provided, in part, by Compute Ontario\footnote{\urlstyle{same}\url{https://www.computeontario.ca}}, Compute Canada\footnote{\urlstyle{same}\url{https://www.computecanada.ca}}, the Province of Ontario, the Government of Canada through CIFAR, and companies sponsoring the Vector Institute\footnote{\urlstyle{same}\url{https://www.vectorinstitute.ai/partners}}. This research is partially funded by The Natural Sciences and Engineering Research Council of Canada (NSERC) through a Discovery Grant to R. E. Mercer. F. Rudzicz is supported by a CIFAR Chair in AI.
\section*{Appendix: A Complete Version of A Data Sample in the Dataset} \label{sec:appendix}

\begin{lstlisting}[language=json]
{"articles":[
    {"PMID":"27976717",
     "TITLE":"Temporal pairwise spike correlations fully capture single-neuron information",
     "ABSTRACT":"To crack the neural code and read out the information neural spikes convey, [...]",
     "INTRO":"Throughout the central nervous system of a mammalian brain [...]", 
     "METHODS":"Deriving the correlation theory of neural information [...]",
     "RESULTS":"We are interested in the information contained in a spike train r(t) about a stimulus s(t)[...]",
     "DISCUSS":"The list of spike timing features that have been implicated in neural coding includes [...]",
     "FIG_CAPTIONS":"Dimensionality of neural information coding [...]",
     "TABLE_CAPTIONS":"Parameter sets across neuron models. [...]",
     "JOURNAL":"Nature communications",
     "YEAR":"2016",
     "DOI":"10.1038/ncomms13805",
     "AUTHORS":[
        "Amadeus,Dettner", 
        "Sabrina,Munzberg", 
        "Tatjana,Tchumatchenko"],
     "MeSH": {
       "D000200":"Action Potentials", 
       "D008959":"Models, Neurological", 
       "D009474":"Neurons", 
       "D059010":"Single-Cell Analysis"
     },
     "CHEMICALS":"None",
     "SUPPLMeSH":"None"
    },
    {
     ...
    },
    ...
]}
\end{lstlisting}

% \nocite{*}
\section{Bibliographical References}\label{reference}
%\label{main:ref}

\bibliographystyle{lrec2022-bib}
\bibliography{lrec2022-example}

\section{Language Resource References}
\label{lr:ref}
\bibliographystylelanguageresource{lrec2022-bib}
\bibliographylanguageresource{languageresource}

\end{document}